\documentclass[sigplan,nonacm]{acmart}
\AtBeginDocument{%
  }

\renewcommand\footnotetextcopyrightpermission[1]{} 
\usepackage{placeins}
\usepackage{needspace}
\begin{document}

\title[AI Telephone Surveying]{AI Telephone Surveying: Automating Quantitative Data Collection with an AI Interviewer}

\author{Danny D. Leybzon}
\authornote{All 3 authors contributed equally and are considered co–primary authors. Results were originally presented at the 2025 American Association of Public Opinion Research (AAPOR) Conference.}
\email{danny@vkl.ai}
\affiliation{%
  \institution{VKL Research, Inc.}
  \city{San Francisco}
  \state{California}
  \country{USA}
}

\author{Shreyas Tirumala}
\authornotemark[1]
\email{shreyas@vkl.ai}
\affiliation{%
  \institution{VKL Research, Inc.}
  \city{San Francisco}
  \state{California}
  \country{USA}
}

\author{Nishant Jain}
\authornotemark[1]
\email{nishant@vkl.ai}
\affiliation{%
  \institution{VKL Research, Inc.}
  \city{San Francisco}
  \state{California}
  \country{USA}
}

\author{Summer Gillen}
\affiliation{%
 \institution{SSRS}
 \city{Glen Mills}
 \state{Pennsylvania}
 \country{USA}}
 
\author{Michael Jackson}
\affiliation{%
 \institution{SSRS}
 \city{Glen Mills}
 \state{Pennsylvania}
 \country{USA}}
 
\author{Cameron McPhee}
\affiliation{%
 \institution{SSRS}
 \city{Glen Mills}
 \state{Pennsylvania}
 \country{USA}}
 
\author{Jennifer Schmidt}
\affiliation{%
 \institution{SSRS}
 \city{Glen Mills}
 \state{Pennsylvania}
 \country{USA}}

\renewcommand{\shortauthors}{Leybzon et al.}

\begin{abstract}
With the rise of voice-enabled artificial intelligence (AI) systems, quantitative survey researchers have access to a new data-collection mode: AI telephone surveying. By using AI to conduct phone interviews, researchers can scale quantitative studies while balancing the dual goals of human-like interactivity and methodological rigor. Unlike earlier efforts that used interactive voice response (IVR) technology to automate these surveys, voice AI enables a more natural and adaptive respondent experience as it is more robust to interruptions, corrections, and other idiosyncrasies of human speech.
\newline
\newline
We built and tested an AI system to conduct quantitative surveys based on large language models (LLM), automatic speech recognition (ASR), and speech synthesis technologies. The system was specifically designed for quantitative research, and strictly adhered to research best practices like question order randomization, answer order randomization, and exact wording.
\newline
\newline
To validate the system’s effectiveness, we deployed it to conduct two pilot surveys with the SSRS Opinion Panel and followed-up with a separate human-administered survey to assess respondent experiences. We measured three key metrics: the survey completion rates, break-off rates, and respondent satisfaction scores. Our results suggest that shorter instruments and more responsive AI interviewers may contribute to improvements across all three metrics studied.
\end{abstract}


\maketitle

\section{Introduction}
Recent advances in artificial intelligence (AI) have created an opportunity to use AI agents to perform telephone surveys. In this paper, we studied the ability of an AI interviewer to conduct structured quantitative telephone surveys with respondents sourced from a probability-based panel of U.S. adults.

To do so, we fielded a 30-minute long survey of 104 members of the SSRS Opinion Panel. This survey instrument had 123 questions, branching logic, and randomization of both question \& answer ordering, which we believe represents the longest survey instrument tested with an AI interviewer to date. By running such a survey, we intend to understand the limits of respondent engagement with AI systems.

\subsection{Background}

Automated voice systems have long been explored as a medium for survey administration both because voice is a faster means of providing input than typing, particularly for “short tasks of a highly interactive nature” \citep{martin1989utility} and because of long-standing research that suggests lower response bias for sensitive questions when human interviewers are not involved \citep{gribble1999interview,cooley2000automating,kreuter2008social}.

Initially, researchers used deterministic Interactive Voice Response (IVR) systems to conduct these types of automated voice surveys. However, research suggested that responses differed from those provided to human interviewers \citep{couper2004does, dillman2009response}, although methods to limit such discrepancies have since been found \citep{johnston2013spoken}. More troublingly, break-off rates\footnote{Refusal Rate 1 (REF1). See \citet{smith2015standard}} for IVR-based surveys have been shown to be considerably higher than traditional computer-assisted telephone interviewing (CATI) surveys, with \citet{gribble2000impact} showing a 24\% IVR break-off rate compared to a 2\% human break-off rate.

In the last three years, automated voice systems have changed considerably. Older statistical methods for natural language understanding have given way to the widespread use of Large Language Models (LLMs), and automated speech recognition systems have since improved too. Recent work has focused on evaluating artificial intelligence (AI) interviewers, systems that leverage automated speech recognition, LLMs, and speech synthesis technologies. Such systems have typically been used to generate more qualitative surveys with intelligently generated follow-up questions \citep{wuttke2024ai, velez2024crowdsourced}.

Contemporaneous work by \citet{lang2025telephone} explored the usage of a modern AI interviewer system to conduct quantitative interviews of college student respondents in Peru. Our collaboration with SSRS instead uses an AI interviewer to administer quantitative surveys to members of the SSRS Opinion Panel, a representative sample of adults across the United States.

\section{Methodology}
\subsection{AI Interviewer}

VKL Research, Inc. developed a voice-agent system designed to conduct telephone surveys with human respondents. The system integrates three key components: real-time automatic speech recognition (ASR), contemporary large language models (LLMs), and speech synthesis.

ASR, commonly known as “speech-to-text,” enables the agent to transcribe spoken responses in real time, mimicking the listening ability of a human interviewer. LLMs provide the AI with verbal reasoning and decision-making capabilities, allowing it to interpret input and select appropriate conversational actions. Finally, speech synthesis technology, also known as “text-to-speech,” is used to vocalize the agent’s responses and instructions to the participant.

The resulting AI interviewer can follow the detailed instructions of a survey instrument, conduct a phone survey conversation, and respond to the varied situations encountered in that context.

\subsection{Respondents}

Respondents were sourced from the SSRS Opinion Panel, a nationally representative probability-based panel of adults aged 18 and older from different geographic locations around the United States. Panelists are recruited using address-based sampling (ABS) and random digit dialing (RDD). All panelists receive monetary incentives for survey completion and decide at the time of recruitment to complete surveys via the web or via telephone. For this study, all data collected was via the telephone in English. 

A sample of 104 potential respondents was randomly selected from the subset of the SSRS Opinion Panel who had chosen telephone as their data collection mode, had previously completed a survey with SSRS, and had not opted out of automated surveys. 

Participants were not informed ahead of time that they would be completing a survey conducted by an AI interviewer. However, during the survey introduction, they were explicitly informed that they were speaking with an AI interviewer and given the opportunity to terminate the survey.

\subsection{Survey Logistics}
Two waves of surveys were conducted with the same panelist pool of 104 potential respondents. These will be referred to as the Wave 1 survey (March 2025) and the Wave 2 survey (April 2025). Both of these surveys were conducted in English.

\textbf{Wave 1 Survey:} The Wave 1 survey was a preliminary test conducted over the course of a weekend from March 7th through March 9th, 2025. All potential respondents were called up to two times each day and given the opportunity to complete the Wave 1 survey instrument. After the Wave 1 survey, targeted technical improvements to the AI interviewer were completed in preparation for the Wave 2 survey.

\textbf{Wave 2 Survey:} The Wave 2 survey was conducted over the course of three weekdays, from April 22nd to April 24th, 2025, roughly six weeks after the Wave 1 Survey. All potential respondents from the sample were called up to two times each day from noon to 9PM local time by the AI interviewer and given the opportunity to complete the Wave 2 survey instrument.  

\textbf{Follow-up Survey:} Several days after each of the Wave 1 and Wave 2 surveys conducted by an AI interviewer, a follow-up survey was fielded. In the follow-up, a human interviewer contacted participants who completed at least one question of the AI-conducted survey, to complete a follow-up survey via telephone about the experience. \newline \newline

\textbf{Additional Considerations:}
\begin{itemize}
  \item Though several questions were shared, the Wave 1 survey used a different instrument from the Wave 2 survey.
  \item The Wave 1 survey was conducted over a weekend in March 2025, from Friday to Sunday, while the Wave 2 survey was conducted during a weekday window in April 2025, from Tuesday to Thursday.
  \item On the first day of the Wave 1 survey, panel members were only called one time and on the second and third days they were called twice. By contrast, for the Wave 2 survey, panel members were called twice each day of the survey.
\end{itemize}

\subsection{Survey Instrument}

\begin{figure*}
  \centering
    \includegraphics[width=.75\linewidth]{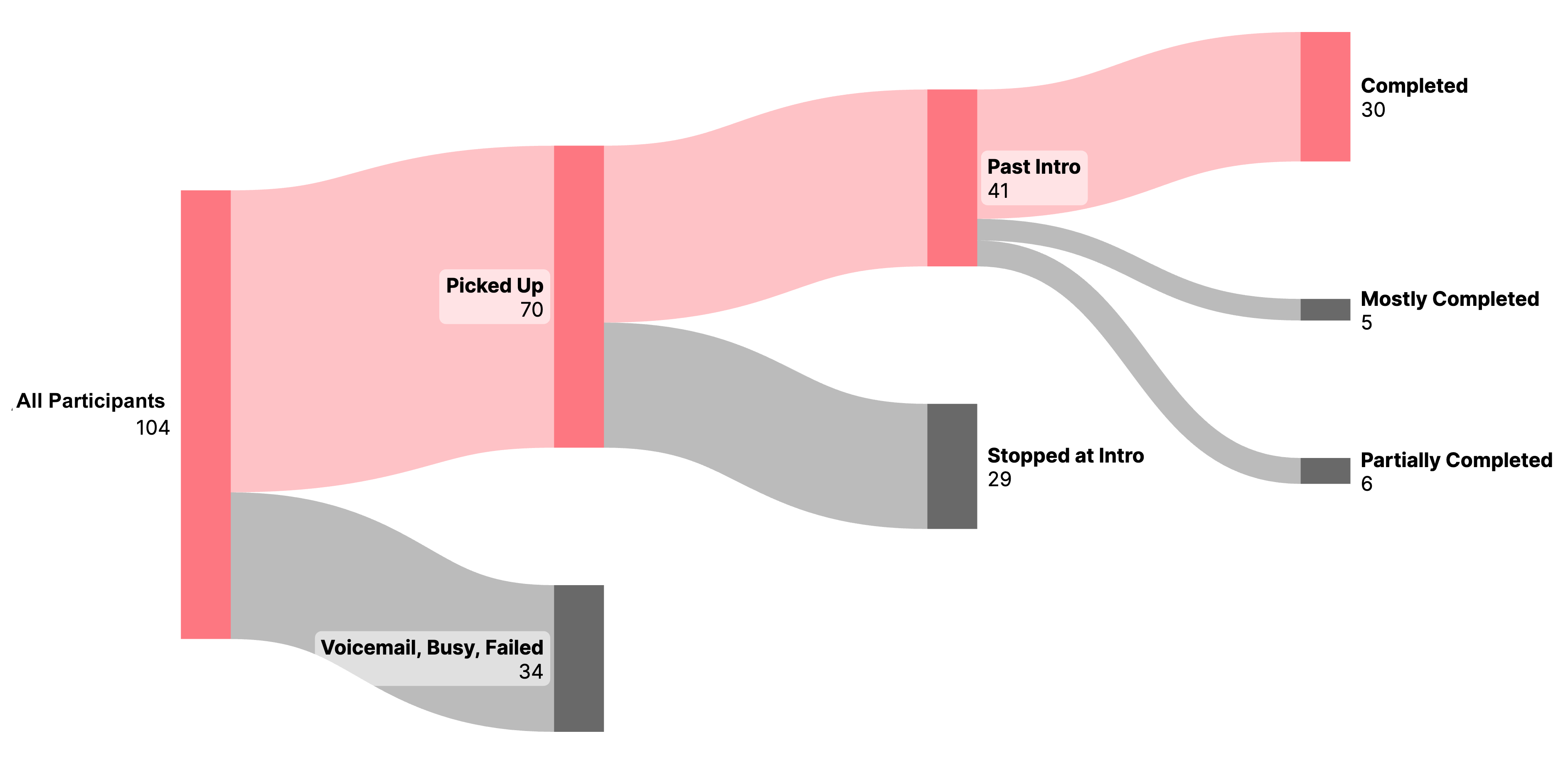}
    \caption{\textbf{Wave 2 Survey Call Dispositions:} A depiction of the dispositions for the 104 potential respondents sampled in Wave 2. Out of the 70 adults sampled who picked up the phone, 30 completed the Wave 2 survey.}
\end{figure*}

\textbf{SSRS Opinion Panel Omnibus:}
The SSRS Opinion Panel Omnibus is a survey fielded twice a month on the SSRS Opinion Panel. It covers a variety of topics across fields, is administered both to web and phone panelists, and contains questions sourced from multiple clients. 

The survey instruments used in this study were actual SSRS Opinion Panel Omnibus surveys, rather than custom-built ones for experimental purposes. While questions from the Omnibus were used to provide a realistic model for a typical human-administered interview, the AI interviews were conducted solely for experimental purposes and were not included in the final Omnibus dataset.

The Wave 2 survey instrument comprised 123 questions of various formats, including single-choice, open-ended, and Likert rating scale items. It featured skip and branching logic, early termination pathways, and multiple layers of randomization. Specifically, subsets of questions were randomized in order, and some questions included randomized answer choices, as dictated by the survey design.

The Wave 1 survey instrument was broadly similar in structure to the Wave 2 survey instrument, but contained different topics and questions.

\textbf{Long and Short Surveys (Wave 2):}
An additional experiment was embedded within the Wave 2 survey to evaluate the impact of survey length on response rates and respondent experience.

Two versions of the instrument were fielded:

\begin{itemize}
    \item A long instrument, containing all 123 questions
    \item A short instrument, containing a subset of 46 questions selected from the long version
\end{itemize}

The short version retained all structural features of the long version, including skip patterns, branching logic, early termination, and randomization of sections, questions, and answer choices. Questions reused in the short version were identical (i.e., no wording changes) to their counterparts in the long version.

Participants were not pre-selected or pre-filtered to receive the long instrument versus the short instrument – random instrument selection happened when calls were initiated.

\section{Results}

We will discuss how two primary factors impact response rates and the respondent experience: AI interviewer quality and questionnaire length. The majority of the discussion will focus on results from the main Wave 2 survey, with relevant information from the preliminary testing conducted in Wave 1 used as necessary.

\begin{figure*}
  \centering
    \includegraphics[width=.75\linewidth]{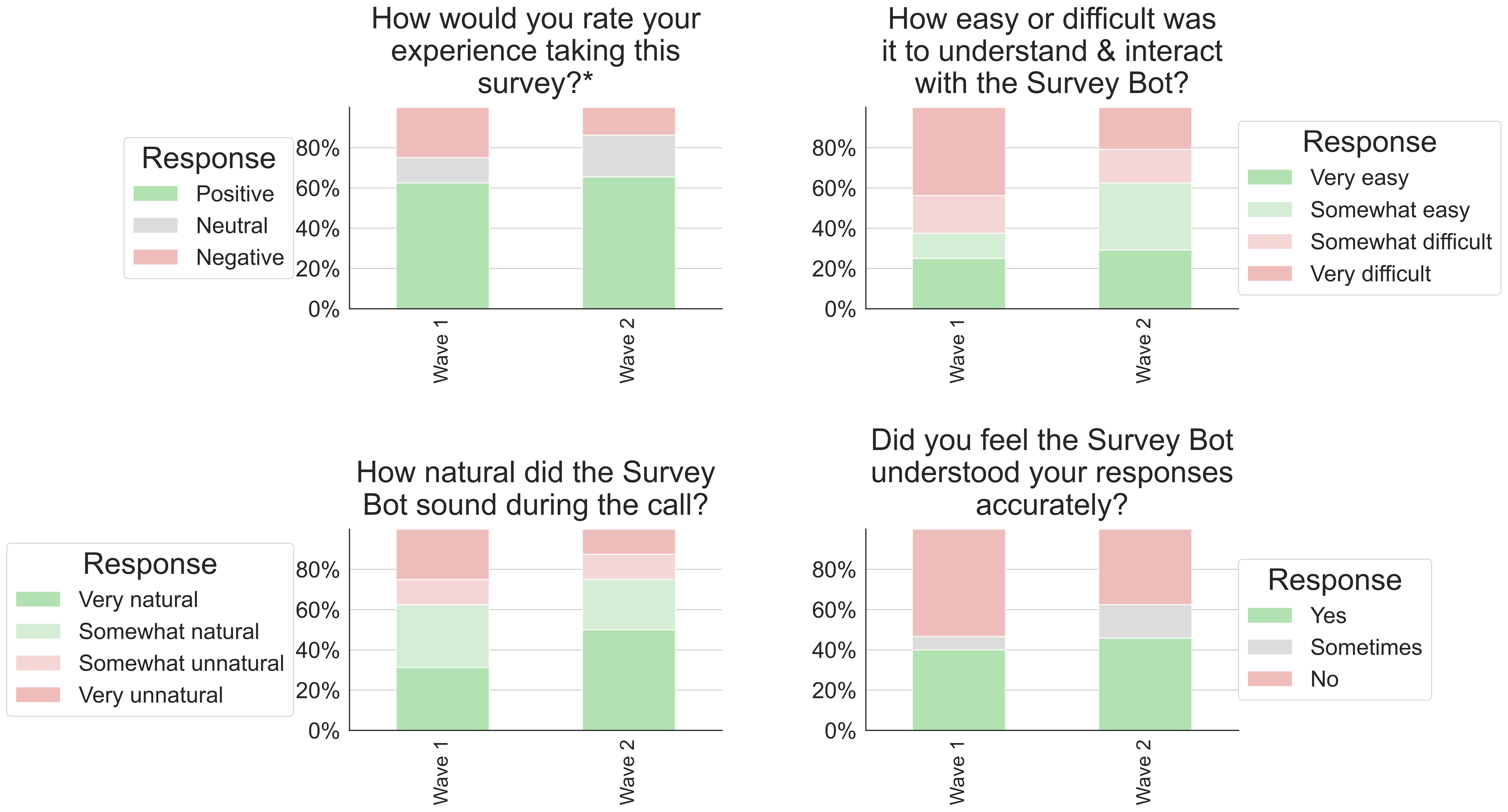}
    \caption{\textbf{Comparison of Respondent Experience Across Waves:} After targeted technical improvements were made to the AI interviewer, respondents in Wave 2 reported having a more positive experience across the board relative to those in Wave 1.}
\end{figure*}

\subsection{Call Flow and Dispositions}

We begin our analysis by dissecting the dispositions of calls conducted during the Wave 2 survey (see Figure 1).

\textbf{Completion Rates:}
Out of 104 adults contacted over the three days, 70 picked up the phone for at least one call. Of these 70 adults, 30 finished the survey, resulting in a completion rate\footnote{Cooperation Rate 1 (COOP1). See \citet{smith2015standard}} of 43\%. 

Of the 70 who picked up the phone, 29 hung up during the introduction (HUDI), resulting in 41 who completed at least one question. This yields a rate of 59\% continuation rate\footnote{Cooperation Rate 2 (COOP2). See \citet{smith2015standard}} past the call introduction. Of the 41 non-HUDI participants, 73\% completed the survey once they had started it (30/41).

\needspace{3\baselineskip}
\textbf{In-survey Experience Ratings:}
Of all the completions in Wave 2, 86\% of respondents classified the experience as neutral or positive in response to an in-survey rating question that was asked at the end of the survey instrument. 21\% percent of respondents rated the survey as excellent, the highest rating available on the scale.

\begin{table}[ht]
\centering
\caption{Response Rates Across Waves}
\begin{tabular}{lcc}
\toprule
 & \textbf{Before (Wave 1)} & \textbf{After (Wave 2)} \\
\midrule
Intro Completion & 42\% & 59\% \\
Survey Completion & 32\% & 73\% \\
\bottomrule
\end{tabular}
\end{table}

\FloatBarrier

\subsection{Impact of AI Interviewer Quality}

We studied the impact of improving the AI interviewer on response rates and respondent experience by comparing the Wave 1 results with the Wave 2 results.

\textbf{AI Interviewer Quality Changes}

After analysis of the Wave 1 survey calls, several qualitative areas for improvement were identified and targeted technical fixes were implemented prior to the Wave 2 survey calls. These are enumerated below.

\begin{itemize}
    \item \textbf{Bot Understanding:} During Wave 1, the AI interviewer occasionally accepted ambiguous responses and made unilateral decisions when mapping them to predefined answer choices. For example, when presented with the options "very liberal" and "somewhat liberal," a respondent might simply reply "liberal." In such cases, the AI would select "somewhat liberal" without further clarification. For Wave 2, the system was modified to proactively probe further when multiple answer choices might apply. 
    
    The AI interviewer also sometimes misheard respondents, particularly for uncommon words. Improvements to real-time speech recognition and comprehension were made to address this.

    \item \textbf{Voice Quality:} At times in Wave 1, the AI interviewer would stutter as it vocalized a question to a respondent. It would also sometimes experience latency, resulting in perceptible delays or silences which sometimes confused respondents. For Wave 2, deployment optimizations were implemented to eliminate stuttering and reduce latency, resulting in a smoother conversation. 

    \item \textbf{Survey Navigation:} We sometimes observed conversational deadlock in Wave 1, where respondents would be waiting for the AI interviewer to say something while the AI interviewer thought it was waiting for the respondent to vocalize an answer. These periods of mutual inaction resulted in confusing periods of silence for respondents. Accordingly, in Wave 2, the AI interviewer was programmed to prompt respondents to get their attention whenever it perceived idleness.
\end{itemize}

\begin{figure*}
  \includegraphics[width=.75\linewidth]{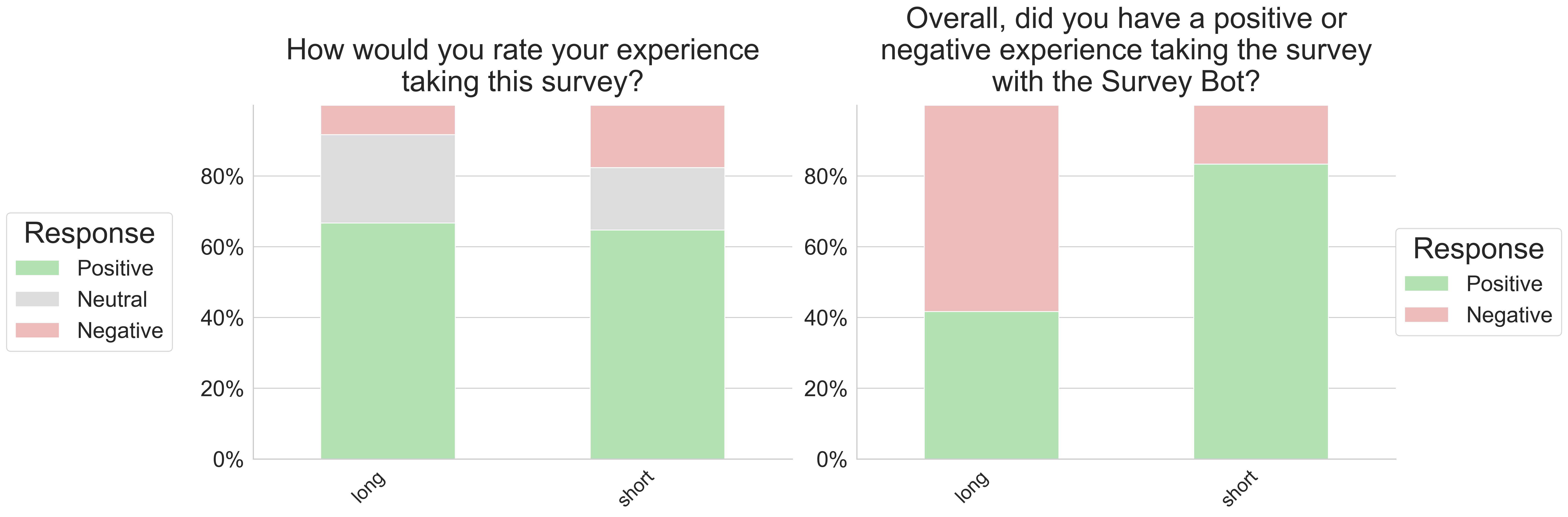}
  \caption{\textbf{Comparison of In-survey Experience and Follow-Up Survey Experience Across Long and Short Survey Cohorts in Wave 2:} While distributions of ratings were similar in the in-survey experience rating question, the follow-up survey results may indicate that shorter surveys are better perceived by respondents.}
\end{figure*}

\begin{figure*}
  \centering
    \includegraphics[width=.75\linewidth]{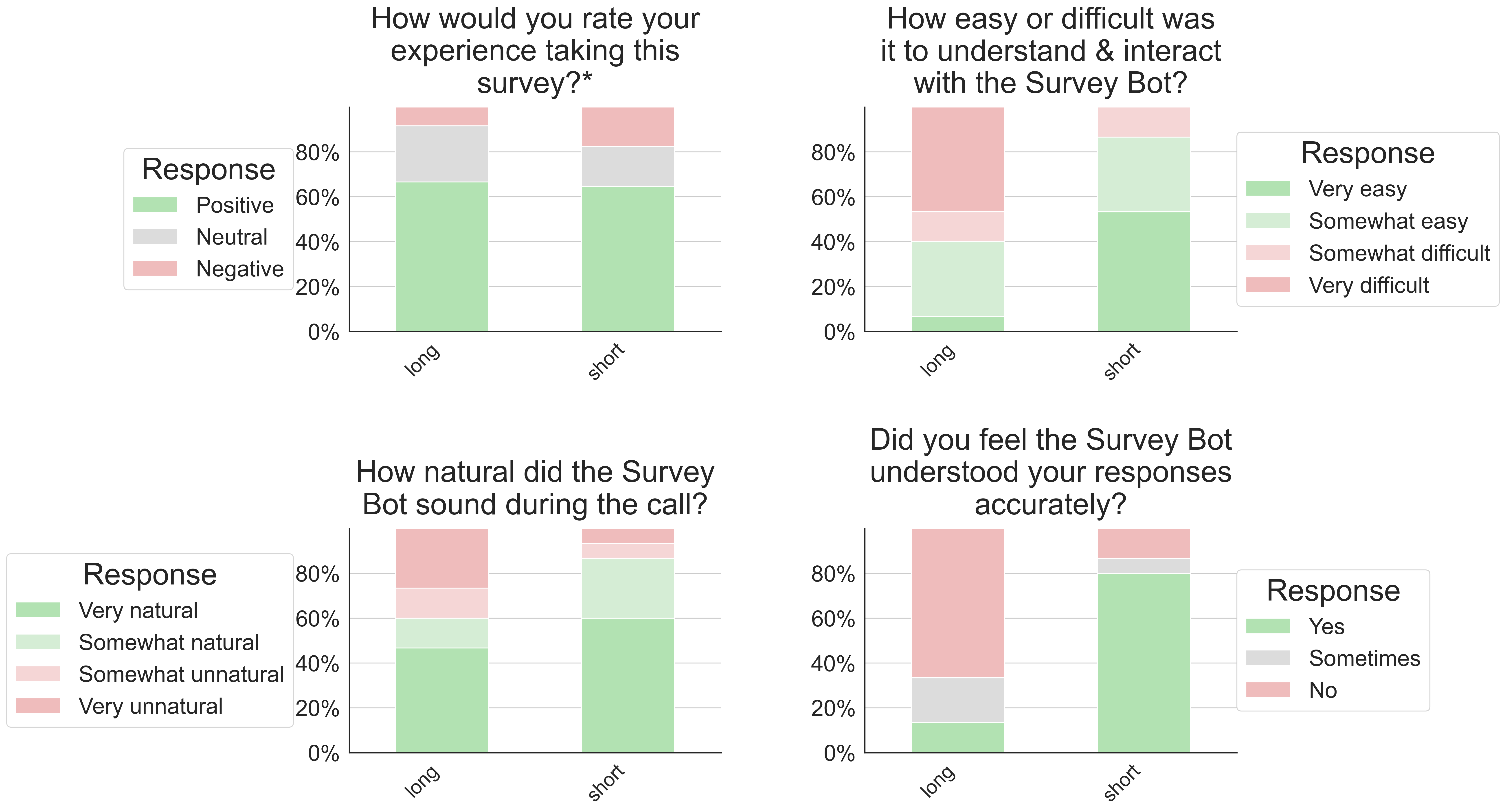}
    \caption{\textbf{Comparison of Respondent Experience Perceptions Across the Short and Long Survey Cohorts in the Follow-up Survey:} Respondents’ evaluations of the experience of taking a survey with the AI interviewer were more positive for those that completed the shorter survey, even though the underlying AI interviewer was the same technology.}
\end{figure*}

\textbf{Response Rate Comparison:}
Comparing response rates between Wave 1 and Wave 2 provides preliminary evidence that AI interviewer quality improvements may have impacted response rates. During Wave 2, we observed a 17 percentage point increase in people who answered the first question after picking up (COOP2 rate per \citet{smith2015standard}) and more than double the percentage of non-HUDI completions (32\% vs 73\%). 


However, these improvements should be interpreted appropriately in context: although AI interviewer quality improvements were made between Wave 1 and Wave 2, the survey instruments and field logistics also differed. In particular, the Wave 2 survey completion rate reported here is a blended rate that includes both long and short surveys, while Wave 1 only fielded long surveys. Non-HUDI completion rates for only the long surveys of Wave 2 were 57\%, which represents a lower, but still substantial 25 percentage point increase.

\textbf{Respondent Feedback Comparison:}
We collected both quantitative and qualitative feedback from respondents to understand how bot behavior impacted their experience. This feedback was gathered via two main methods:

\begin{itemize}
    \item An in-survey question asked at the end of the AI-administered survey.
    \item A follow-up telephone survey conducted several days later by a human interviewer.
\end{itemize}

It is important to note that participation differed between the AI-administered survey and the follow-up surveys. Not all respondents who completed the AI-administered survey were successfully reached for the follow-up, and some who participated in the follow-up had only partially completed the original survey.

Across both Wave 1 and Wave 2, the vast majority of respondents (75\% in Wave 1, 86\% in Wave 2) reported a neutral to positive experience based on an in-survey feedback question.

In the Wave 2 follow-up survey (n = 24), the distributions for all questions related to the AI interviewer’s characteristics (e.g. level of understanding, interaction quality, etc.) were more positive than in the follow-up survey for Wave 1.


In the qualitative feedback provided to human CATI interviewers during the follow-up survey, Wave 2 respondents noted that the AI interviewer felt more natural and was better at understanding them. They also reported enjoying the novelty of the experience and the ease of interaction with the AI.

\subsection{Impact of Survey Length}

While the previous section focused on the differences between the Wave 1 and Wave 2 surveys, this next section focuses solely on the Wave 2 survey. As previously mentioned, we fielded two instruments in the Wave 2 survey:

\begin{itemize}
    \item A long instrument containing 123 questions.
    \item A short instrument containing a 46-question subset of the long survey.
\end{itemize}

We refer to these as the long survey and the short survey respectively. Both instruments contained skip and branching logic, termination scenarios, and randomization at the question and answer choice levels.

At the end of each survey, respondents were asked to rate their experience. Among those who completed the survey, in-survey experience ratings were similar between long and short survey groups.

However, when asked about their experiences in a follow-up survey, a far higher proportion of respondents reported a positive experience with the short survey than the long.\footnote{Note that sample sizes were different across the in-survey ratings and the follow-up survey since not every respondent to the AI-administered survey picked up the phone for the follow-up survey.}

Notably, in the follow-up survey, respondents who received the short survey reported more positive evaluations of the interview than those who received the long version. This trend was consistent across multiple questions related to perceptions of the AI interviewer's interactions, its ability to understand respondents, and naturalness of the AI interviewer’s speech. Importantly, both survey versions were administered by the same AI interviewer, and a manual audit of call recordings revealed no substantial differences in the AI's behavior between the two cohorts: the primary distinction was survey length. This may indicate that survey length impacts subjective perceptions of experience with AI interviewers.

\section{Strengths of AI Interviewing}

Through this study, we observed several situations in which AI interviewers may present advantages over traditional interactive voice response (IVR) systems. 

\subsection{Handling Ambiguity}

LLMs enable AI interviewers to conduct limited verbal reasoning; this allows them to manage ambiguous or unexpected responses more effectively than previous automated systems. For instance, when a respondent indicated “don’t know” in response to a specific question, the AI interviewer was able to gently probe them once more to make their best attempt at answering the question. Moreover, after technical improvements made for Wave 2, when a respondent gave vague responses (e.g., “liberal” to a question with options such as “somewhat liberal” and “very liberal”), the system detected the ambiguity and proactively asked for clarification. These capabilities reflect a level of verbal understanding not present in most current IVR systems.

\subsection{Robustness to Respondent Behavior}

Surveys sometimes involve unexpected interactions unrelated to answering questions: respondents may interrupt, pause, or momentarily step away from the conversation. Unlike IVR systems, which often fail in such cases, AI interviewers can appropriately respond to these behaviors. In one notable example, a respondent asked for a moment to find a quieter location. The AI interviewer waited patiently and resumed the conversation, ultimately enabling the respondent to complete the survey without issue.

\subsection{Dealing with Audio Issues}

Real-world survey calls frequently take place in noisy environments with ambient noise that can challenge IVR. The AI interviewer demonstrated an impressive ability to handle challenging conditions, including background noise (e.g., crying children, traffic, television) and muffled or accented speech. During manual audits, there were several instances where human reviewers had difficulty understanding a respondent’s audio, but the AI interviewer had correctly interpreted the response and continued.

\section{Challenges with AI Interviewing}

\subsection{Transcription Issues}
Real-time transcription of what a respondent says during a survey conversation is still not one hundred percent accurate. Even state-of-the-art automatic speech recognition (ASR) systems can miss syllables, sounds, or words. Since live transcriptions from ASR systems are the primary input for an AI interviewer’s decision-making, any transcription errors can disrupt the flow of the survey and require respondents to repeat themselves, potentially leading to frustration or drop-off. AI transcription errors, if not caught and corrected, may also become a source of measurement error. Additional research with larger sample sizes will be helpful to assess the potential impact of AI transcription errors and how transcription error rates compare to analogous errors with human and IVR interviewers.

\begin{figure*}
  \centering
    \includegraphics[width=.75\linewidth]{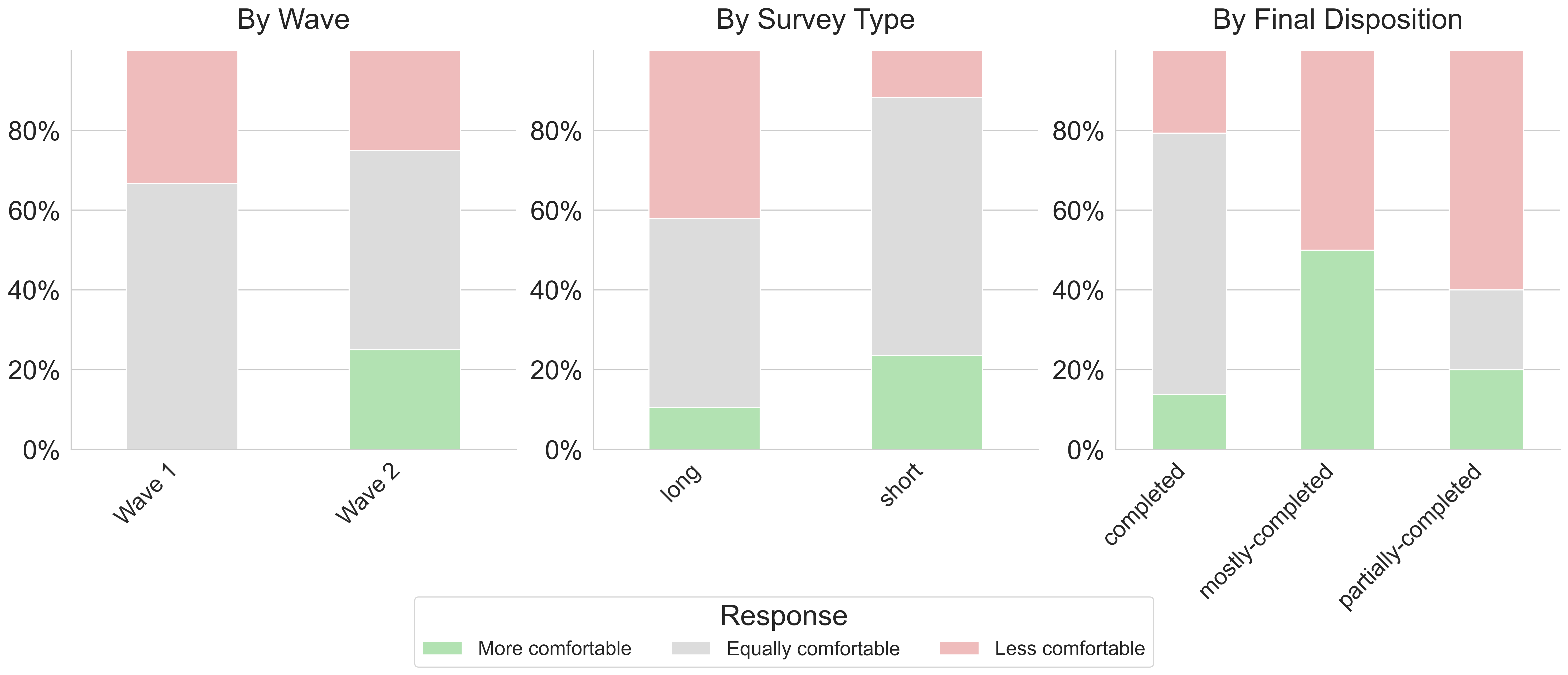}
    \caption{\textbf{Comparison of Respondent Comfort with AI Interviewers Relative to Human Interviewers:} Some respondents reported more comfort with an AI interviewer than a human interviewer. A majority of respondents in both waves reported at least equal comfort with an AI interviewer. The short survey cohort indicated higher levels of comfort with AI interviewers relative to human interviewers.}
\end{figure*}

\subsection{Strictness and Flexibility}
Balancing survey standardization with the flexibility of a natural-sounding conversation can pose challenges; interviewers must make judgment calls about when to accept responses and when to re-prompt respondents. If an AI system relies on extremely strict criteria for acceptable responses, respondents may feel frustrated and hang up. However, overly permissive acceptance criteria may compromise data quality. Striking the right balance often takes time and requires case-by-case calibration of AI systems on a per-survey basis, according to the goals of each survey designer.

\subsection{Respondent Misbehavior}
We observed a couple of instances of straightlining, where respondents attempted to provide the same response to a series of questions without listening to the questions in their entirety. While human interviewers are often trained to both identify straightlining and respond by stopping the survey, the version of the AI interviewer used in this study did not include any logic to detect and mitigate such behavior. Incorporating such detection is an important area for future development.

We hypothesize that the lack of social accountability associated with speaking with a human may cause some respondents to feel more comfortable straightlining with an AI interviewer. While we cannot confirm this hypothesis based on our current data, we encourage further research into this topic.

\section{Respondent Comfort with AI Interviewers}

\subsection{Sensitive Topics and Comfort with AI Interviewers}
We are particularly intrigued by the potential efficacy of AI interviewers to conduct interviews that contain sensitive topics. Both the Wave 1 survey and the Wave 2 survey included questions about sensitive topics like personal finances, crime, and violence. Previous work suggests that, for sensitive topics, modes that have direct interactions with humans may be less preferred than modes that afford more privacy, such as IVR or self-administered web surveys \citep{kreuter2008social,goldstein2025digging}.

The early results observed in follow-up surveys in this work appear to corroborate these findings. Most respondents self-reported to be at least equally comfortable speaking with an AI interviewer or with a human interviewer about these topics (see Figure 5).

\subsection{AI Interviewer Quality and Respondent Comfort}

In both the Wave 1 and Wave 2 surveys, a majority of respondents reported equal comfort or more comfort with AI interviewers. In Wave 2, six respondents reported more comfort speaking with an AI interviewer than a human, an increase from zero in Wave 1. Understanding when and why respondents indicate more comfort with an AI interviewer requires further study. 

\subsection{Survey Length and Respondent Comfort}

Based on follow-up survey results for Wave 2, survey length may have a relationship to a respondent’s level of comfort with an AI interviewer. Compared to respondents given the long survey, a higher proportion of respondents given the short survey expressed feeling equally or more comfortable with an AI interviewer.

An important caveat, however, is that the number of study participants who completed the follow-up survey was small (n=24), so further research is needed. 

\section{Conclusion}
We find that AI interviewers show significant promise for conducting structured, quantitative telephone surveys. The AI interviewer was able to administer a real-world SSRS Opinion Panel Omnibus Survey of representative length (123 questions / ~30 minutes) and complexity. Among respondents who completed the survey, 86\% reported a neutral or positive experience, and some respondents reported feeling greater comfort speaking with an AI interviewer than a human interviewer.

Our findings also indicate that AI interviewer quality and survey length may have an effect on both response rates and the overall respondent experience. Technical improvements made to the AI interviewer between Wave 1 and Wave 2 coincided with increases in both completion and satisfaction metrics. Additionally, respondents who received shorter survey instruments were more likely to report positive experiences in follow-ups, despite the underlying AI system used to administer both instruments remaining the same.

Qualitative observations of AI-administered survey calls suggest that the AI interviewer has strengths in handling ambiguity, remaining robust to respondent behaviors, and dealing with audio quality issues. However, AI interviewers still have challenges with maintaining 100\% accurate real-time transcription, balancing strictness and flexibility, and responding to respondent misbehavior. 

As AI interviewer technology continues to evolve, these systems may play a role in enabling scalable, cost-effective, and respondent-friendly survey data collection.

\section{Future Work}
While these results highlight the promise of AI interviewers in survey research, further inquiry is needed across several areas.

\subsection{Larger-Scale Validation}
104 adults from the SSRS Opinion Panel participated in this study, with 70 participants answering at least one call. 30 respondents completed the survey in Wave 2. Conducting similar experiments with larger numbers of participants may allow for more robust conclusions about respondent behavior, preferences, and data quality (including mode effects on estimates).

\subsection{Respondent Preferences and Demographics}
Several open questions remain about how different populations engage with AI interviewers:
\begin{itemize}
    \item Which types of respondents prefer speaking to an AI interviewer when being surveyed?
    \item Are there particular contexts or topics for which respondents prefer an AI interviewer over a human interviewer?
    \item Are the types of respondents who complete surveys with an AI interviewer demographically representative of the broader U.S. population?
    \item Do findings based on incentivized panelists generalize to random digit dialing (RDD)-based samples?
\end{itemize}

Questions about demographic representativeness are particularly salient given the increased use of mixed-mode designs for probabilistic samples. In such designs, phone interviews are commonly used as a complement to web-based efforts with the explicit aim of improving representation of demographics that are underrepresented in online surveys. While the sample sizes for this initial pilot did not allow for a conclusive comparison of respondent demographics, it will be important to ascertain whether AI-enabled interviews are similarly effective at bringing in underrepresented demographics. 

\subsection{Inbound Surveying}
While this study looked exclusively at outbound calling, AI interviewers can also be used in an inbound context. Since these AI interviewers can be available at any time of day to conduct a survey, doing so may afford further scheduling flexibility for both respondents and survey researchers. In the inbound calling scenario, the same key questions explored in this study about AI interviewer performance, response rates, and respondent experience are promising lines of inquiry for future research.



\bibliographystyle{ACM-Reference-Format}
\bibliography{sample-base}


\end{document}